\documentclass[10pt,twocolumn,letterpaper]{article}

\usepackage{iccv}
\usepackage{times}
\usepackage{epsfig}
\usepackage{graphicx}
\usepackage{amsmath}
\usepackage{amssymb}

\iccvfinalcopy 

\begin{document}


\title{FH-GAN: Face Hallucination and Recognition using Generative Adversarial Network}

\author{Bayram Bayramli\qquad Usman Ali\qquad Te Qi\ \qquad Hongtao Lu\\
Shanghai Jiao Tong University\\
\tt\small\{bayram\_bai, usmanali, qite1030, htlu\}@sjtu.edu.cn
}

\maketitle

\begin{abstract}
There are many factors affecting visual face recognition, such as low resolution images, aging, illumination and pose variance, etc. One of the most important problem is low resolution face images which can result in bad performance on face recognition. Most of the general face recognition algorithms  usually assume a sufficient resolution for the face images. However, in practice many applications often do not have sufficient image resolutions. The modern face hallucination models demonstrate reasonable performance to reconstruct high-resolution images from its corresponding low resolution images.  However, they do not consider identity level information during hallucination which directly affects results of the recognition of low resolution faces. To address this issue, we propose a Face Hallucination Generative Adversarial Network (FH-GAN) which improves the quality of low resolution face images and accurately recognize those low quality images. Concretely, we make the following contributions: 1) we propose FH-GAN network, an end-to-end system, that improves both face hallucination and face recognition simultaneously. The novelty of this proposed network depends on incorporating identity information in a GAN-based face hallucination algorithm via combining a face recognition network for identity preserving. 2) We also propose a new face hallucination network, namely Dense Sparse Network (DSNet), which improves upon the state-of-art in face hallucination. 3) We demonstrate benefits of training the face recognition and GAN-based DSNet jointly by reporting good result on face hallucination and recognition. 

\end{abstract}


\section{Introduction}

In recent years, super-resolution models \cite{Chao1}, \cite{Wang},  \cite{Jimmy} which produce high-resolution (HR) images from low-resolution (LR) images has progressed tremendously thanks to the deep learning techniques. Since it is an ill posed problem, LR input may correspond to many HR candidate images which may lead to losing identity information. Many existing works do not consider identity information while hallucinating LR face images, as a result they cannot produce HR faces similar to the real identity. On the other hand, the extensive use of surveillance systems and security cameras makes a challenging use case for face recognition in an environment where detected faces will be in low resolution. Although some face recognition methods \cite{Yaniv1}, \cite{Yaniv2}, \cite{YiSun1}, \cite{YiSun2} achieved satisfactory results, these algorithms cannot perform well on the low resolution images. Since LR face images may match with many HR candidates, this uncertainty may lead to distorted identity information. Based on these facts, we can see that recovering identity information can improve low resolution face recognition systems and as well as performance of face hallucination. 

\begin{figure}[t]
\center
\includegraphics[width=5 cm, height= 3 cm]{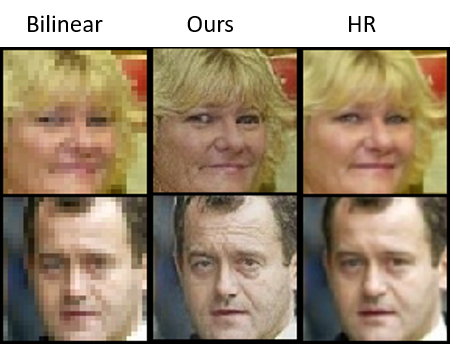}
\caption{Hallucination example of our method.}
\label{fig:fron_fig}
\end{figure}

To address this issue, we aim to answer how to hallucinate low resolution face images which can also improve face recognition performance. The goal of the proposed method, FH-GAN, is to enhance upon the visual quality and recognizabilty of low resolution facial images by considering the identity information recovery during super-resolution process. The architecture of FH-GAN is illustrated in Figure 2.

Specifically, we propose an end-to-end FH-GAN network to hallucinate low resolution faces and preserve the identity information which is qualified for face recognition. To achieve it, we introduce:
\begin{figure*}
\begin{center}
\includegraphics[width=13 cm]{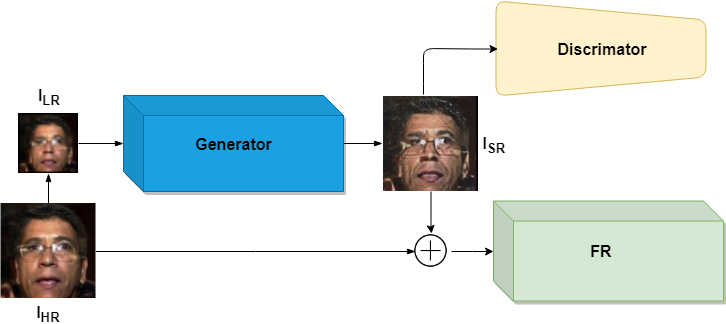}
\end{center}
\caption{The architecture of our proposed FH-GAN consists of three associated networks: 1) the main network as well as the generator network is a newly proposed Face Hallucination network(sub-section 3.1). 2) discriminator network used to distinguish between HR face image and hallucinated face image(see sub-section 3.2). The third network is Face Recognition network for recognizing on the hallucinated face images and enhancing face hallucination through an identity loss(see sub-section 3.3). FR - Face Recognition and $\oplus$ denotes concatenation.}
\label{fig:main_architecture}
\end{figure*}

\begin{itemize}
    \item a novel generator architecture for GAN which is sparsely aggregating the output of previous layers at any given depth. 
    It offers fewer parameters, improves flow of information through the network and alleviates gradient vanishing problem.
    
    \item our GAN-based face hallucination utilizes both pixel level and feature level information as the supervisory signal to preserve the identity information.
    
    \item identity loss which measures identity difference between hallucinated HR image and ground truth HR images by using  the face recognition.
\end{itemize}

\section{Related Work}
In this section, we review the related work in image super-resolution, face hallucination, and face recognition.

\textbf{Single Image Super Resolution (SISR)}. SISR aims to reconstruct HR image from its corresponding LR input.  Many super resolution methods have been developed including classical approaches \cite{classic1}, \cite{classic2}, \cite{classic3} and deep learning based approaches \cite{dp1}, \cite{dp2}. In recent years, huge improvements in deep learning methods have also resulted in significant enhancements in image super resolution techniques. The first work that utilized convolutional networks for super-resolution purposes was SRCNN by Dogn et al., \cite{first_cnn} to predict mapping between interpolated LR and HR pair images using three layers of convolutional networks. This benchmark was further enhanced by expanding network depth. To further improve the reconstruction accuracy \cite{dp2}, \cite{rel_w} used more convolutional deep neural networks. They both used interpolation of original LR images as an input which causes an increase in computation and information loss. Later on, \cite{dp1} used sub-pixel convolutional layer to learn effective upscaling. Notably, we also use sub-pixel layer in our network. Later, \cite{srgan} exploited advantage of residual learning by using sub-pixel layer. However, all these methods ignore to take advantages of information from each convolutional layer. Consequently, these methods lose useful hierarchical features from LR image. \cite{srdense} introduced the basic dense block from DenseNet \cite{densenet} to learn hierarchical features but the problem with this method is that feature maps aggregated by dense skip connections are not fully exploited. To solve these issues, we propose sparsely aggregated skip connection blocks in our generator network (DSNet) to concatenate features at different levels.

\textbf{Face Hallucination}. Image SR methods can be applied to all kind of images which do not incorporate face-specific information. Generally, face hallucination is a type of class-specific image SR. \cite{fh_rl} introduced bichannel convolutional networks to hallucinate face images in the wild. \cite{fh_rl1} introduced two-step auto-encoder architecture to hallucinate unaligned, noisy low resolution face images.\cite{sicnn} also introduced identity information recovery in their proposed method. \cite{fh_gan1} proposed GAN-based method to super resolve very low resolution image without using perceptual loss. Except from \cite{sicnn} which is not using GAN-based generator, above mentioned methods do not consider identity information in hallucination process which is vital for recognition and visual quality. In our method, we used perceptual loss to achieve more realistic results and identity loss to incorporate with face recognition model to facilitate identity space by utilizing advanced GAN method. Our experiments demonstrate indistinguishable visual quality images and improve the performance of low resolution face recognition.
\begin{figure*}
\begin{center}
\includegraphics[width=15 cm]{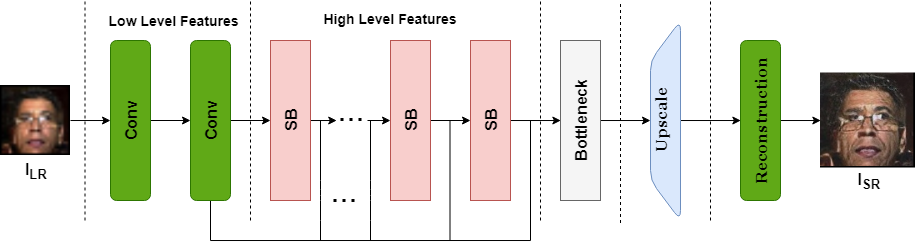}
\end{center}
\caption{The architecture of our proposed super-resolution network, DSNet.}
\label{fig:dsnet}
\end{figure*}

\textbf{Face Recognition}. The low-resolution face recognition task is a subset of the face recognition. There are many useful application scenarios for this task such as security cameras and surveillance systems. In this scenario, face images are captured in the wild from cameras with a large standoff. Some state-of-art techniques \cite{face_recg1}, \cite{face_recg2}, \cite{face_recg3} has already achieved an accuracy over 99 percent. However, those algorithms can only deal effectively on faces with large region of interest. Therefore, when resolution drops, the performance of these algorithms drops respectively. \cite{vlr} proposed a relationship-learning-based SR between the high-resolution image space and the LR image space. \cite{face_recg_study} showed the problem of very low resolution recognition cases through deep learning based architecture. 

This is one of the main motivations in our work. We employed the face recognition model of \cite{arcface}. ArcFace model provides excellent performance on face verification on high resolution images as shown in \cite{arcface}. In our paper, ArcFace is trained specifically to preserve identity of low resolution face image as well as to enhance the face image quality while hallucinating. As a result, one of our contributions is to demonstrate that a face recognition model when incorporated and trained end-to-end with a super resolution network can still give high accuracy on low resolution face images.

\section{Method}
In this section, we will first describe the proposed architecture including three connected networks and their loss functions: the first network is a super-resolution network which is also used as a generator, Densely connected Sparse Blocks network (DSNet),  used to super-resolve LR face images to HR face images. The second one is an adversarial network used to distinguish super-resolved images from HR correspond. The third network is Face Recognition for identity preserving on the hallucinated facial images. In the end we will describe our identity loss. During evaluation time, the discriminator is not used. In general, we call our algorithm FH-GAN, shown in Fig.2

\subsection{Face Hallucination Network}
Notably, we propose an architecture that aims to learn end-to-end mapping function between low-resolution facial image $I_{LR}$ and it's corresponding high-resolution facial images $I_{HR}$. As shown in figure 3, Dense Sparse network (DSNet) is mainly composed of four parts: low level feature extractor(LLFE), sparely aggregated CNN blocks for learning high level features (SparseBlock - SpB), upscaling layer for increasing the resolution size and a reconstruction layer for generating the HR output. 

\textbf{LLFE.} We denote $I_{LR}$ and $I_{SR}$ as the input and output of DSNet. Specifically, we use two convolutional layers, from now on we call Conv, to extract shallow level features. The first Conv layer extracts features from LR input
\begin{equation}
    y_0 = F_{LLFE_0}(I_{LR}),
\end{equation}
where $F_{LLFE_0}(\cdot)$ denotes the convolution operation and $y_0$ is the output of first low level feature extractor. The output of (1) will be the input of second Conv layer
\begin{equation}
    y_1 = F_{LLFE_1}(y_0),
\end{equation}
where $F_{LLFE_1}(\cdot)$ denotes the second low level feature extractor convolution operation and $y_1$ is the output of respective layer.

\textbf{Sparse Blocks (SpB).} After applying LLFE layers to learn low level features, $(2)$ is used as input to Sparse Blocks for learning high-level features. The sparse block structure is inspired by sparse aggregation in convolutional networks, first proposed in \cite{sparsenet}. In the structure of SparseNet \cite{sparsenet} feature maps from previous layers are sparsely concatenated together rather than directly summed as in ResNets \cite{resnet}. As shown in Figure 4, each sparse block in our network consists of multiple layers, where each layer is a composition of a convolution followed by PReLu activation function. Within a sparse block, rather than concatenating features from all previous layers, the number of incoming links to a layer are reduced by aggregating the state of preceding  layers at an exponential offsets; for example $i-1, i - 2, i - 4, i - 8...$ layers will be concatenated as input for $i$-th layer.  The output of $l$-th convolutional layer in SpB is computed as:
\begin{equation}
    y_l = \sigma(W_l[y_{l-c^0}, y_{l-c^1}, y_{l-c^2},....,y_{l-c^k}])
\end{equation}
where $[y_{l-c^0}, y_{l-c^1}, y_{l-c^2},....,y_{l-c^k}]$ refers to the concatenation of feature maps, $W_l$ is the weights of the $l-th$ Conv layer and $\sigma$ denotes the PReLU [] activation function. Bias term is omitted for simplicity. $c$ is a positive integer and $k$ is the largest non-negative integer such that $c^k \leq l$.

The main difference of SparseNet from DenseNet and ResNet is that the input to a particular layer is formed by aggregation of a subset of previous outputs. The power of short gradient paths is maintained in the Sparse Blocks. The importance of short paths is to enhance the flow of information thence alleviating the vanishing gradient problem. Moreover, altering the number of incoming links to be logarithmic, the sparse block architecture drastically reduce the number of parameters, thereby require less memory and computation cost to achieve high performance.

Multiple sparse blocks are joined together to constitute a high-level feature learner component. Each sparse block receives a concatenation of low-level features from (2) and all preceding sparse blocks as input via skip connections. This enables each sparse block to directly see low-level as well as high-level feature information for better reconstruction performance.   

\textbf{Bottleneck layer}. As described above, features from the previous SpB are introduced directly to the next SpB in a concatenation way. This  yields a large sized input for the subsequent up-sampling layer, so it is essential to reduce the features size. It has been studied in \cite{memnet} that a convolutinonal layer size of 1 x 1 kernel can be utilized as a bottleneck layer to diminish the size features map. To enhance model computational efficiency, we utilize bottleneck layer to diminish number of features before feeding them to upsampling layer. The number of feature maps is reduced to 128.

\textbf{UpSampling and Reconstruction layer}. We use sub-pxiel \cite{dp1} to upscale the LR feature maps to HR feature maps. The ultimate Conv layer in the DSNet which has 3 x 3 kernel size and 3 channels is used for reconstruction.
\subsubsection{Pixel and perceptual loss}
Given a set of low resolution images $I_{LR}$ and its corresponding high resolution images $I_{HR}$ we minimize the Mean Squared Error(MSE) in image space which is named Pixel-wise loss:
\begin{equation}
    l_{pixel} = \frac{1}{N}\sum_{i=1}^{N}||I_{HR}^i  - G(I_{LR})^i||^2
\end{equation}
where $G(\cdot)$ represents the output of generator network and $N$ is the batch size. Although, MSE loss achieves high PSNR values, it usually results in blurry and unrealistic images. To handle this, perceptual loss is proposed in \cite{percp_loss} to achieve visually good and sharper images. In perceptual loss, MSE is used in feature space of hallucinated image and its corresponding HR image. We extracted features of HR image and hallucinated image from VGG-19 \cite{vgg19} to calculate the following loss:
\begin{equation}
    l_{perceptual} = \frac{1}{N}\sum_{i=1}^{N}||\phi(I_{HR}^i)  - \phi(G(I_{LR})^i)||^2
\end{equation}
where $\phi$ denotes the feature maps obtained from the last convolutional layer of VGG-19[] and $G(I_{LR})^i$ is the $i-th$ super-resolved face image.

\begin{figure}
\begin{center}
\includegraphics[width=8.5 cm]{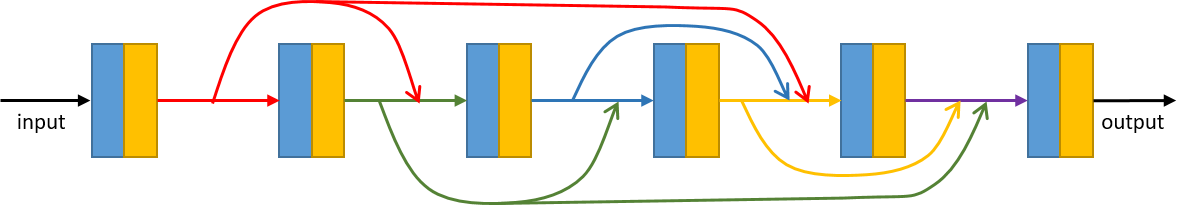}
\end{center}
\caption{The architecture of our Sparse Block.}
\label{fig:sparse_block}
\end{figure}

\subsection{Adversarial Network}
In this subsection, we define adversarial loss to produce realistic super resolved face images. The idea of using GAN \cite{gan} is straightforward: the goal of discriminator D is to distinguish super-resolved images generated by generator G from the original images. The generator G aims to generate realistic face images to fool D. In DSNet, we use Wasserstein GAN (WGAN)\cite{wgan} which is then improved in WGAN-GP \cite{wgan-gp}. The reason to use WGAN-GP is not to enhance the quality of hallucinated face images but to stabilize and reduce the overall training time. As the generator of WGAN-GP we use our super-resolution network and for the discriminator network we utilize the discriminator of DCGAN \cite{dcgan} without using batch normalization.

\textbf{Adversarial Loss.} We employ the WGAN-GP loss in our face hallucination network:
\begin{equation}
\begin{aligned}
            l_{WGAN} = \mathbb{E}\underset{\hat{I} \sim \mathbb{P}_g}{}[D(\hat{I})] - \mathbb{E}\underset{I \sim \mathbb{P}_r}{}[D(I^{HR})] \\
    + \lambda \mathbb{E}\underset{\hat{I} \sim \mathbb{P}_{\hat{I}}}{}[(||\nabla_{\hat{I}}D(\hat{I})||_2 -1)^2],
\end{aligned}
\end{equation}

where $\mathbb{P}_r$ is the input data distribution and $\mathbb{P}_g$ is the generator $G$ distribution defined by $\hat{I} = G(I_{LR})$ is obtained by uniformly sampling along straight lines between pairs of samples from $\mathbb{P}_r$ and $\mathbb{P}_g$. $\lambda$ is a penalty coefficient which we set to 10 in our experiments.

\subsection{Face Recognition Network}
Herein, we employ ArcFace as our face recognition model due to it is state-of-the-art performance on identity representation. ArcFace is Resnet-like \cite{resnet} CNN model and it is trained by Additive Angular Margin Loss(ArcFace) which can effectively enhance the disciminative power of feature embeddings. ArcFace loss function is modified traditional Softmax loss. The keypoint in ArcFace is that the classification boundary is maximized directly in the angular space. More details about ArcFace can be found here \cite{arcface}. The loss function of ArcFace on a training image sample is represented as:
\begin{equation}
    l_{ArcFace}(y_i) = - \frac{1}{N}\sum_{i=1}^N\log\frac{e^{s(cos(\theta_{y_i} + m))}}{e^{s(cos(\theta_{y_i} + m)) + \sum_{j=1, j\neq y_i}^{n}e^{s cos\theta_j}}}
\end{equation}
where $y_i$ is the $i$-th sample, N is a batch size. $m$ is the hyperparameter of angular margin and $s$ is the feature scale.
Given a mini-batch, we compute the $l_{ArcFace}$ on concatenation of non-paired $I_{HR}$ and $I_SR$ face images. We train ArcFace using the following loss:
\begin{equation}
    l_{FR} = l_{ArcFace}(\{I_{HR}^i, I_{SR}^i\})
\end{equation}
where, \{ \} denotes concatenation.
\subsubsection{Identity loss}
Equation (4), (5), (6) have been used in general purpose super-resolution. Although, they do provide decent results for facial super-resolution, during the super-resolution process identity information is easy to be lost as these losses are not incorporating information related to face identity information. We have examined that when these losses are used alone identity details may be missing and the performance of the face recognition decreasing (see Table 3.)

To alleviate this issue, we propose to enforce facial identity consistency between the low and the high resolution face images via integrating face recognition network. Simply, we further use a constrain on the identity level.  Therefore, for better preservation of human face identity of the super-resolved images, identity-wise feature representation with face recognition network used as supervisory signal. The identity loss described as follows:
\begin{equation}
    l_{identity} =  \frac{1}{N}\sum_{i=1}^{N}||FR(I_{HR}^i)  - FR(G(I_{LR})^i)||^2
\end{equation}
where $FR(I_{HR}^i)$ and $FR(G(I_{LR})^i)$ are the identity features extracted from the fully connected layer of our face recognition model. $G(I_{LR})^i$  represents $i$-th generated facial images.
\subsection{Overall training loss}
In summary, the overall losses used for training FH-GAN is weighted sum of the above loss functions:
\begin{equation}
    l_{total} = \lambda_1l_{pixel} + \lambda_2l_{perceptual} + \lambda_3l_{WGAN} + \lambda_4l_{id}
\end{equation}
where $\lambda_1$,  $\lambda_2$, $\lambda_3$, $\lambda_4$ are the corresponding loss weights.
\section{Experiments}
In this section,  description of the training and testing details will be first provided followed by the implementation details. Afterwards, we will discuss the comparisons with others and benefit of our method. Later, we will present the effectiveness of using identity loss. Furthermore, we report standard super-resolution metrics, PSNR and SSIM, of proposed FH-GAN. According to \cite{srgan}, the result of PSNR and SSIM are not indicative of visual quality. To alleviate the issue with poor metrics of PSNR and SSIM, we also propose an indirect way to evaluate face image super-resolution quality based on face recognition result.  We report face verification accuracies on different methods. In particular, we trained ArchFace on high resolution and hallucinated face images and then used it for verification of face images on low-resolution images.

\subsection{Experimental Settings}
\textbf{Dataset.} VGGFACE2 \cite{data_vgg} is a large-scale dataset for face recognition and synthesis which cover a large range of pose, age and ethnicity. A total of 9000 identities contain images from a wide range of different ethnicities, accents, professions and ages. We use 8631 identities of 3.31 million images for training face recognition model. To train face hallucination, we randomly select 1.2M images from VGGFACE2 dataset.

We use two different  datasets for our proposed method. The first one is LFW \cite{data_lfw} dataset used for testing both face verification and face hallucination performance in the wild. The LFW contains 13,233 images from 5,749 identities. We use CFP \cite{data_cfp} dataset to evaluate face verification. The CFP contains 7000 images from 500 identities. These two dataset are considered in unconstrained settings. Several state-of-the-art models such as, SRGAN \cite{srgan}, SRDenseNet \cite{srdense}, RDN \cite{rdn} have been used to compare our approach.

\textbf{Data Preprocessing.} In order to conduct a fair comparison with other methods, training data is detected by MTCNN \cite{mtcnn} and aligned to a canonical view of size 112~x~112. 

\textbf{Implementation details.} HR image size was cropped and aligned to 112x112 and LR input image was obtained by downsampling the HR images using bilinear kernel with a scale factor of 4x. 

To train ArcFace,  we employed ResNet34 \cite{resnet} and set the embedding features to 512. We follow \cite{resnet} to set the feature scale $s$ to 64 and choose the angular margin $m$ of ArcFace at 0.5 We set the batch size to 256 and the learning rate is started from 0.01 and divided by 10 after 15, 18 epochs. The training process has finished at 20 epoch.

To train GAN-based DSNet, we used 6 Sparse Blocks while each Sparse Block has 6 convolutional layers. In total, depth of face hallucination network size is 41 layers including, sparse blocks, low level feature extractors, bottleneck, upsampling and reconstitution layers. Within each Sparse Block, we used growth rate of 32. Low level feature extractors have filter size of 64 and size of all convolutional layers were set to 3x3 except bottleneck layer, where size is 1x1. The parametric rectified linear units (PReLu) was used as  the activation function. All the networks were optimized using Adam. We used the mini batch size of 128. The learning rate is set to 1e-3 and gradually decreased to 1e-5. Training has finished at 56k iterations. 

For end-to-end training of the FH-GAN, all networks(DSNet, discriminator and ArcFace) were training jointly for 4 epochs and learning rate of 1e-4. Face Hallucination model and ArchFace were trained using Adam~\cite{adam} and SGD respectively. All models are implemented in PyTorch.

\subsection{Discussions}
We compare our method to the other methods, including, SRDenseNet \cite{srdense}, RDN \cite{rdn}, SRGAN \cite{srgan} to demonstrate the effectiveness of our proposed method.

\textbf{Difference to SRDenseNet.} First and foremost, SRDenseNet uses local dense connections from DenseNet \cite{densenet} which concatenates all the outputs of previous layers thus results in over-burdening the model. However, concatenation allows every subsequent layer a clean view of all previous features but densely concatenation of features mean that a primary portion of the model is dedicated to process previously seen features. Consequently, it is hard for the model to make full use of dense skip connections and all the parameters. But, we exploit the local sparse connections into our proposed network inspired from SparseNets \cite{sparsenet} which concatenated the features in an logarithmic manner rather than a linear manner. This property allows to utilize larger growth rate, which is filter size, and enlarge our model by using more layers. By using sparse aggregation topolopy in our proposed method, we reduce parameters size to half and achieve faster convergence compare to SRDenseNet. Another difference is that SRDenseNet only uses MSE loss but we use multiple losses to make the model robust to get better hallucinated face images. As a result, our method achieves better performance and generate visually pleasing face images.

\textbf{Difference to SRGAN and RDN}. In terms different choice of loss function, we mainly summarize differences of our method compared with SRGAN and RDN. RDN only uses $L_1$ loss function but in contrast we do not only use pixel level information but we also incorporate feature level information in our method. Using only pixel-wise loss will result in blurry images and lose identity information which is very crucial for face recognition. However, SRGAN utilizes feature level loss (perceptual loss) to make super-resolved images sharper but sometimes super-resolved images have some artifacts, such as white and red spots on the face. Additionally, SRGAN does not consider to preserve identity information in metric space which will lead to miss identity information and generate additional artifacts in super-resolved images. In our method, we use perceptual loss as well as identity loss  to impose identity level constraint by jointly training face hallucination model with face recognition model.

\textbf{Benefit of our model.} In summary, by using sparse blocks we can further increase our model size and growth rate which is very beneficial for super-resolution task to use very deep networks in two aspects 1) large amount of contextual information can be utilized from LR images; 2) in very deep networks high nonlinearity generated by PReLu layers can be utilized to model the sophisticated mapping functions between LR and HR. By using sparse blocks, we get better flexibility and parameter efficiency. As can be seen in Fig. 5, our method provides the sharper and more detailed results performing well across different kind of face images.

\subsection{Effectiveness of Identity Loss}
\textbf{Identity Loss.} Table 1 shows the ablation investigation on the effects of identity loss. We find that, face recognition performance decreases when we do not include identity loss in our propose method. As we said earlier, because of ill-posed behavior of face hallucination methods it is easier to lose identity information during hallucination. 

As shown in Table 1, we get better accuracy when we train FH-GAN jointly with face recognition network. We constrain identity level information by adding face recognition loss. The Identity level difference can be measured by robust face recognition model.  The face recognition model with the identity-wise feature representation is used as supervisory signal which helps to preserve identity information and increase the performance of face verification. 
\begin{table}
\begin{center}
\begin{tabular}{|l|c|c|}
\hline
Method & Identity Loss & Accuracy \\
\hline
FH-GAN & x & 99.00 \% \\
\hline
FH-GAN & \checkmark & \textbf{99.14 \%} \\
\hline
\end{tabular}
\end{center}
\caption{Effectiveness of identity loss on Face Verification Performance.}
\end{table}

\subsection{Super Resolution Results}

\begin{figure*}
\begin{center}
\includegraphics[width=15 cm]{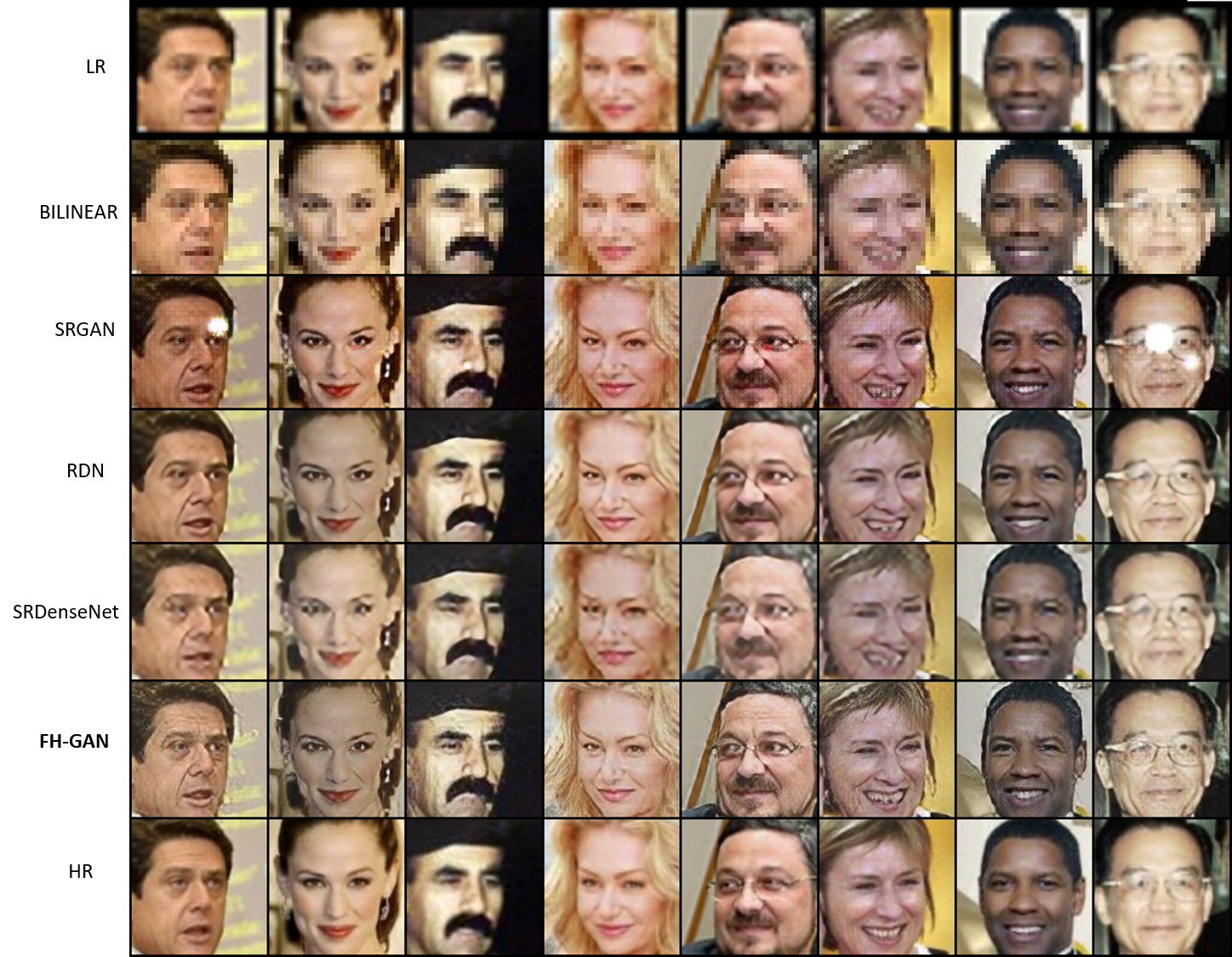}
\end{center}
\caption{From top to bottom: LR image, Bilinear interpolation, SRGAN\cite{srgan}, \cite{rdn}, \cite{srdense}, our model and HR}
\label{fig:images_compr}
\end{figure*}

 We compared the PSNR and SSIM results using the proposed method and using other state-of-the-art super-resolution methods, including bilinear interpolation. As we discussed, because of robustness of our model, it achieves better results as compared to others. In most of the cases, standard metrics, such as PSNR and SSIM,  for super resolution are not very reliable for visually better images.
\begin{table}
\begin{center}
\begin{tabular}{|l|c|c|}
\hline
Method & LFW ACC& CFP ACC \\
\hline
FR-Bilinear & 98.62 \% & 92.3 \% \\
\hline
FR-SRGAN & 99.03 \% & 93.08 \%  \\
\hline
FR-RDN & 98.92 \% & 92.6 \% \\
\hline
FR-SrDenseNet & 98.87 \% & 92.16 \% \\
\hline
FH-GAN & \textbf{99.16 \%} & \textbf{93.36 \%} \\
\hline
FR-HR images & 99.47 \% & 95.05 \% \\
\hline
\end{tabular}
\end{center}
\caption{Face verification results on LFW and CFP dataset. FR stands for Face Recognition model which we used in our all experiments. The results in this case, are indicative of visual quality. FR-Bilinear means this method super-resolved the face image using bilinear interpolation and run Face Recognition model on that and similarly other methods.}
\end{table}
 
\begin{table}
\begin{center}
\begin{tabular}{l|c|c}
\hline
Method & PSNR & SSIM \\
\hline
Bilinear upsample & 20.3 & 0.76\\
\hline
SR-GAN & 20.78 & 0.77\\
\hline
SRDenseNet & 20.26 & 0.79\\
\hline
RDN & 21.26 & 0.81\\
\hline
Ours & \textbf{21.35} & \textbf{0.83}\\
\hline
\end{tabular}
\end{center}
\caption{PSNR and SSIM based Face Hallucination performance on LFW. The results are not indicative of visual quality.}
\label{tab:psnr}
\end{table}

 Although bilinear method is fast and very light in super resolving but the face images generated by this method are blurry and have artifacts. Bilinear method fails to super resolve low resolution images. Face images generated by RDN and SRDenseNet result in over-smoothed images because of learning only pixel-wise information. Consequently, over-smoothed images do not contain face features completely.  As shown in Fig. 5, SRGAN faces contains white dots artifacts in hallucinated face images. Because of effectiveness of our generator network and identity loss we comparatively obtain visually good images.
 
 A few failure cases of our method can be seen in Fig.~6. These failure cases are primarily because of large occlusions and multiple faces. 
 In these failure cases our super-resolved images still preserve the identity but are distorted. Improving these images and investigating real low quality images are left for future works.

\subsection{Face Recognition Results}
The proposed FH-GAN aims to recognize low resolution human faces. Therefore, for verifying the identity preserving capacity of different super-resoution models, face recognition on two benchmark datasets is studied. We evaluate the performance of face verification on LFW dataset and CFP dataset by using the ArcFace extracted features of hallucinated face images.

\textbf{Face Verification on low resolution LFW and CFP.} Face verification performance evaluated on the recognition accuracy (ACC) in the wild are shown in Table 3. From the results RDN and SRDenseNet are flawed because of their weak specificity to identity preservation. Even though SRGAN has utilized perceptual loss but still their face verification accuracy is not good because they do not consider identity preservation in identity metric space. Our model achieves best results of face verification on two datasets which are very close to face verification results on HR face images. This is indicative of superiority of our face hallucination method.

\begin{figure}[t]
\center
\includegraphics[width=7 cm, height= 3 cm]{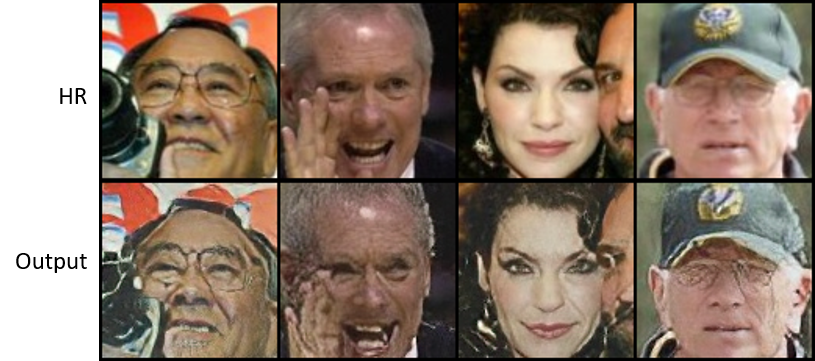}
\caption{Hallucinated examples of visually bad results produced by our method. These images include large occlusions.}
\label{fig:failure_case}
\end{figure}

\section{Conclusion}
This paper has answered how to hallucinate and recognize the faces simultaneously if the face image resolution is not sufficient enough. Specifically, we proposed FH-GAN: an end-to-end system for super-resolving face images and recognizing those images. Our method incorporates facial identity information in a newly proposed generator architecture using WGAN for face hallucination. The face recognition model aims to improve identity preservation and quality of hallucinated images. We show improvements on both face hallucination and low resolution face recognition.

{\small
\bibliographystyle{ieee}
\bibliography{ms.bbl}
}

\end{document}